\documentclass{article} 
\usepackage{iclr2016_conference,times}
\usepackage{url}
\usepackage{graphicx}
\usepackage{amssymb,amsmath,amsthm,amsfonts}
\usepackage{algorithm}
\newtheorem{theorem}{Theorem}

\title{Invariant backpropagation: how to train \\
a transformation-invariant neural network}

\author{Sergey Demyanov \thanks{http://www.demyanov.net} \\
IBM Research Australia, Melbourne, VIC, Australia\\
\texttt{sergeyde@au1.ibm.com} \\
\AND
James Bailey, Ramamohanarao Kotagiri, Christopher Leckie \\
Department of Computing and Information Systems \\
The University of Melbourne, Parkville, VIC, Australia, 3010 \\
\texttt{\{baileyj, kotagiri, caleckie\}@unimelb.edu.au} \\
}

\begin{document}

\maketitle

\begin{abstract}
In many classification problems a classifier should be robust to small variations in the input vector. This is a desired property not only for particular transformations, such as translation and rotation in image classification problems, but also for all others for which the change is small enough to retain the object perceptually indistinguishable. We propose two extensions of the backpropagation algorithm that train a neural network to be robust to variations in the feature vector. While the first of them enforces robustness of the loss function to all variations, the second method trains the predictions to be robust to a particular variation which changes the loss function the most. The second methods demonstrates better results, but is slightly slower. We analytically compare the proposed algorithm with two the most similar approaches (Tangent BP and Adversarial Training), and propose their fast versions. In the experimental part we perform comparison of all algorithms in terms of classification accuracy and robustness to noise on MNIST and CIFAR-10 datasets. Additionally we analyze how the performance of the proposed algorithm depends on the dataset size and data augmentation.

\end{abstract}

\section{Introduction}

Neural networks are widely used in machine learning. For example, they are showing the best results in image classification (\cite{szegedy2014going, lee2014deeply}), image labeling (\cite{karpathy2014deep}) and speech recognition. Deep neural networks applied to large datasets can automatically learn from a huge number of features, that allow them to represent very complex relations between raw input data and output classes. However, it also means that deep neural networks can suffer from overfitting, and different regularization techniques are crucially important for good performance. 

It is often the case that there exist a number of variations of a given object that preserve its label. For example, image labels are usually invariant to small variations in their location on the image, size, angle, brightness, etc. In the area of voice recognition the result has to be invariant to the speech tone, speed and accent. Moreover, the predictions should always be robust to random noise. However, this knowledge is not incorporated in the learning process. 

In this work we propose two methods of achieving local invariance by extending the standard backpropagation algorithm. First of them enforces robustness of the loss function to all variations in the input vector. Second methods trains the predictions to be robust to variation of the input vector in the direction which changes the loss function the most. We refer to them as Loss Invariant BackPropagation (Loss IBP), and Prediction IBP. While one of them is faster, the other one demonstrates better performance. Both methods can be applied to all types of neural networks in combination with any other regularization technique. 

\subsection{Backpropagation algorithm}

We denote $K$ as the number of layers in a neural network and $y_i, \; i \in \{0,\ldots,K\}$ as the activation vectors of each layer. The activation of the first layer $y_0$ is the input vector $x$. If the input is an image that consists of one or more feature maps, we still consider it as a vector by traversing the maps and concatenating them together. The transformation between layers might be different: convolution, matrix multiplication, non-linear transformation, etc. We assume that $y_{i} = f_i(y_{i-1}; w_i)$, where $w_i$ is the set of weights, which may be empty. The computation of the layer activations is the first (forward) pass of the backpropagation algorithm. Moreover, the loss function $L(y_K)$ can also be considered as a layer $y_{K+1}$ of the length $1$. The forward pass is thus a calculation of the composition of functions $f_{K+1}(f_K(\ldots f_1(x) \ldots))$, applied to the input vector $x$.

Let us denote the vectors of derivatives with respect to layer values $\partial L / \partial y_i$ as $dy_i$. Then, similar to the forward propagating functions $y_i = f_i(y_{i-1}; w_i)$, we can define backward propagating functions $dy_{i-1} = \tilde{f}_i(dy_i; w_i)$. We refer to them as \textit{reverse} functions. According to the chain rule, we can obtain their matrix form:
\begin{equation} \label{eq:jacobian}
dy_{i-1} = \tilde{f}_i(dy_i; w_i) = dy_i \cdot J_i(y_{i-1}; w_i),
\end{equation}
where $J_i(y_{i-1})$ is the Jacobian, i.e. the matrix of the derivatives $\partial y_i^j / \partial y_{i-1}^k$. The backward pass is thus a consecutive matrix multiplication of the Jacobians $\prod_{i=K+1}^1 J_i(y_{i-1})$ of layer functions $f_i(y_{i-1}; w_i)$, computed at the points $y_{i-1}$. Note, that the first Jacobian $J_{K+1}(y_K)$ is the vector of derivatives $dy_K = \partial L / \partial y_K$ of the loss function $L$ with respect to predictions $y_K$. The last vector $dy_0 = \prod_{i=K+1}^1 J_i(y_{i-1}) = \nabla_x L$ contains the derivatives of the loss function with respect to the input vector.

Next, let us also denote the vector of weight gradients $\partial L / \partial w_i$ as $dw_i$. Then we can write the chain rule for $dw_i$ in a matrix form as $dw_i = J_i^w(y_{i-1}; w_i) \cdot dy_i$, where $J_i^w(y_{i-1}; w_i)$ is the Jacobian matrix of the derivatives with respect to weights $\partial y_i^j / \partial w_i^{kl}$. However, if $f_i$ is a linear function, the Jacobian $J_i^w(y_{i-1}; w_i)$ is equivalent to the vector $y_{i-1}^T$, so
\begin{equation} \label{eq:weights_der}
dw_i = y_{i-1}^T \cdot dy_i
\end{equation}
In this article we consider all layers with weights to be linear.

After the $dw_i$ are computed, the weights are updated: $w_i \leftarrow w_i - \alpha \cdot dw_i, \; \forall i \in \{1,\ldots,K\}, \; \alpha > 0$. Here $\alpha$ is the coefficient that specifies the size of the step in the opposite direction to the derivative, which usually reduces over time.

\section{Related work}

A number of techniques that allow to achieve robustness to particular variations have been proposed. Convolutional neural networks, which consist of pairs of convolutional and subsampling layers, are the most commonly used one. They provide robustness to small shifts and scaling, and also significantly reduce the number of training parameters compared to fully-connected perceptrons. However, they are not able to deal with other types of variations. Another popular method is data augmentation. It assumes training on the objects, artificially generated from the existing training set using the transformation functions. Unfortunately, such generation is not always possible. There exist two other approaches, which also attempt to solve this problem analytically using the gradients of the loss function with respect to input. We discuss them below.



\subsection{Tangent backpropagation algorithm} \label{sec:tangent}
The first approach is Tangent backpropagation algorithm (\cite{simard2012transformation}), which allows to train a network, robust to a set of \textit{predefined} transformations. The authors consider some invariant transformation function $g(x; \theta), \, s.t. \, g(x, 0) = x$, which must preserve the predictions $p(g(x; \theta))$ within a local neighborhood of $\theta=0$. Since the predictions $p(x)$ in this neighborhood must also be constant, a necessary condition for the network is
\[
\left. \nabla_\theta p(g(x,\theta)) \right|_{\theta=0} = 0
\]
To achieve this, the authors add a loss regularization term $R(x)$ to the main loss function $L$:
\begin{equation} \label{eq:tbpfun}
L_{min}(x) = L(p(x)) + \beta R(x) = L(p(x)) + \beta \tilde{L}(\nabla_\theta p(g(x,\theta))|_{\theta=0}), \,\,\, \tilde{L}(z) = \frac{1}{r}||z||_r^r
\end{equation}
Using the chain rule we can get obtain the following representation for $\nabla_\theta p(g(x,\theta))|_{\theta=0}$:
\[
\nabla_\theta p(g(x,\theta))|_{\theta=0} = \nabla_x p(x) \cdot \left. \nabla_\theta g(x; \theta) \right|_{\theta=0} = \prod_{i=K}^1 J_i(y_{i-1}) \cdot \left. \nabla_\theta g(x; \theta) \right|_{\theta=0}
\]
The last term depends only on the function $g(x; \theta)$ and the input value $x$, and therefore can be computed in advance. The authors refer to $\nabla_\theta x = \left. \nabla_\theta g(x; \theta) \right|_{\theta=0}$ as \textit{tangent vectors}. 
The authors propose to compute the additional loss term by initializing the network with a tangent vector $\nabla_\theta x^T$ and propagating it through a \textit{linearized} network, i.e., consecutively multiplying it on the transposed Jacobians $J_i^T(y_{i-1})_{i=\{1,\ldots,K\}}$. Indeed, 
\[
\nabla_\theta x^T \cdot \prod_{i=1}^{K} J_i^T(y_{i-1}) = \prod_{i=K}^1 J_i(y_{i-1}) \cdot \nabla_\theta x = \nabla_x p(x) \cdot \nabla_\theta x = \nabla_\theta p(g(x,\theta))|_{\theta=0}
\]


The main drawback of Tangent BP is computational complexity. As it can be seen from the definition, it linearly depends on the number of transformations the classifier learns to be invariant to. The authors describe an example of training a network for image classification, which is robust to five transformations: two translations, two scalings, and rotation. In this case the required learning time is 6 times larger than for the standard BP.

The usage of tangent vectors also makes Tangent BP more difficult to implement. To achieve this, the authors suggest to obtain a continuous image representation by applying a Gaussian filter, which requires additional preprocessing and one more hyperparameter (filter smoothness). While the basic transformation operators are given by simple Lie operators, other transformations may require additional coding.

\subsection{Adversarial Training} \label{sec:adversarial}


The second algorithm is a recently proposed Adversarial Training (\cite{goodfellow2014explaining}). In (\cite{szegedy2013intriguing}) the authors described an interesting phenomena: it is possible to artificially generate an image indistinguishable from the image of the dataset, such that a trained network's prediction about it is completely wrong. Of course, people never make such kinds of mistakes. These objects were called \textit{adversarial examples}. In (\cite{goodfellow2014explaining}) the authors showed that it is possible to generate adversarial examples by moving into the direction given by the loss function gradient $\nabla_x L(p(x))$, i.e.,
\begin{equation} \label{eq:advex}
x^*(x;\epsilon) = x + \epsilon \, sign(\nabla_x L(p(x)))
\end{equation}
In a high dimensional space even a small move may significantly change the loss function $L(p(x))$. To deal with the problem of adversarial examples, the authors propose the algorithm of \textit{Adversarial Training} (AT). The idea of the algorithm is to additionally train the network on the adversarial examples, which can be quickly generated using the gradients $\nabla_x L(p(x))$, obtained in the end of the backward pass. Adversarial Training uses the same labels $l(x)$ for the new object $x^*$ as for the original object $x$, so the loss function $L(p(x^*(x;\epsilon)))$ is the same. The updated loss function is thus
\begin{equation} \label{eq:atorigloss}
L_{min}(x) = (L(p(x)) + L(p(x^*(x;\epsilon))))/2
\end{equation}



Adversarial training is quite similar to the Tangent propagation algorithm, but differs in a couple of aspects. First, Adversarial training uses the gradients of the loss function $\nabla_x L(p(x))$, while Tangent BP uses tangent vectors $\nabla_\theta x^T$. Second, while Adversarial Training propagates the new \textbf{objects} $x^*$ through the \textbf{original} network, Tangent BP propagates the \textbf{gradients} $\nabla_\theta x^T$ through the \textbf{linearized} network. The proposed Prediction IBP algorithm can be also derived by combining these properties.

\section{Invariant backpropagation}

In the first part of this section we describe \textit{Loss} IBP, which makes the main loss function robust to all variations in the input vector. In the second part we describe \textit{Prediction} IBP, which aims to make the network predictions robust to the variation in the direction specified by $\nabla_x L(p(x))$. While both versions use the gradients $\nabla_x L(p(x))$, they differ in their loss functions, computational complexity, and also in experimental results.


\subsection{Loss IBP} \label{sec:fastIBP}

In many classification problems we have a large number of features. Formally it means that the input vectors $y_0$ come from a high dimensional vector space. In this space every vector can move in a huge number of directions, but most of them should not change the vector's label. The goal of the algorithm is to make a classifier robust to such variations.

Let us consider a $K$-layer neural network with an input $x = y_0$, and predictions $p(x) = y_K$. Using the vector of true labels $l(x)$, we compute the loss function $L(p(x)) = y_{K+1}$, and at the end of the backward pass of backpropagation algorithm we obtain the vector of its gradients $dy_0 = \nabla_x L(p(x)) = \prod_{i=K+1}^1 J_i(y_{i-1})$. This vector defines the direction that changes the loss function $L(p(x))$, and its length specifies how large this change is. In the small neighborhood we can assume that $dL \approx dx \cdot \nabla_x L^T(p(x))$. If $\nabla_x L(p(x))$ is small, then the same change of $x$, will cause a smaller change of $L$. Thus, a smaller vector length corresponds to a more robust the classifier, and vice versa. Let us specify the additional loss function
\begin{equation} \label{eq:ibp:fastloss}
\tilde{L}_r(\nabla_x L(p(x))) = \tilde{L}_r(dy_0) = \frac{1}{r}||dy_0||_r^r, \, d\tilde{y}_0 = \frac{\partial \tilde{L}_r(dy_0)}{\partial (dy_0)}
\end{equation} 
which is computed at the end of the backward pass. In order to achieve robustness to variations, we need to make it as small as possible. By default we assume $r=1$.

Note that $\tilde{L}_2(dy_0)$ is very similar to the Frobenius norm of the Jacobian matrix, which is used as a regularization term in contractive autoencoders (\cite{rifai2011contractive}). The minimization of $\tilde{L}(dy_0)$ encourages the classifier to be invariant to changes of the input vector in all directions, not only those that are known to be invariant. At the same time, the minimization of $L(p(x))$ ensures that the predictions change when we move towards the samples of a different class, so the classifier is not invariant in these directions. The combination of these two loss functions aims to ensure good performance. In order to minimize the joint loss function
\begin{equation} \label{eq:ibp:jointloss}
L_{min}(x) = L(p(x)) + \beta \tilde{L}(\nabla_x L(p(x))),
\end{equation}
we need to additionally obtain the derivatives of the additional loss function with respect to the weights $d\tilde{w}_i = \nabla_{w_i} \tilde{L}(dy_0)$. In Section \ref{sec:lossimpl} we discuss how to efficiently compute them, using only one additional forward pass. Once these derivatives are computed, we can update the weights using the new rule
\begin{equation} \label{eq:ibp:invupdaterule}
  w_i \leftarrow w_i - \alpha (dw_i + \beta \cdot d\tilde{w}_i) \; \alpha \geq 0, \beta \geq 0,
\end{equation}
Here $\beta$ is the coefficient that controls the strength of regularization, and plays a crucial role in achieving good performance. Note that when $\beta = 0$, the algorithm is equivalent to the standard backpropagation. Since the additional loss function aims to minimize the gradients of the main loss function $\nabla_x L(p(x))$, we call this algorithm Loss IBP.

\subsection{Prediction IBP}

While Loss IBP makes the main loss function $L(p(x))$ robust to variations, it does not necessarily imply the robustness of the predictions $p(x)$ themselves. Unfortunately we cannot compute the gradients of predictions with respect to the input vector as their dimensionality can be very large. However, we can compute the gradients of predictions in the direction given by $\nabla_x L(p(x))$. As it was shown in Section \ref{sec:adversarial}, movement in this direction can generate adversarial examples, whose predictions significantly differ from $x$. We can thus introduce another additional loss function
\begin{equation} \label{eq:ibp:slowloss}
\tilde{L}_r(\nabla_\epsilon p(x + \epsilon \nabla_x L(p(x)))|_{\epsilon=0}) = \tilde{L}_r(\nabla_x p(x) \cdot \nabla_x L^T(p(x)))
\end{equation} 
We call the algorithm with this loss function Prediction IBP. The only difference of Prediction IBP from Tangent BP is the initial vector for the third pass. While Tangent BP uses precomputed tangent vectors, Prediction IBP uses the vector of gradients $\nabla_x L(p(x))$, obtained at the end of the backward pass. The weight gradients of the additional loss function $\tilde{L}$ can be computed the same way as they are computed in Tangent BP. Therefore, Prediction IBP always requires two times more computation time than standard BP. 

\begin{figure}[h]
\centering
\includegraphics[height=4cm]{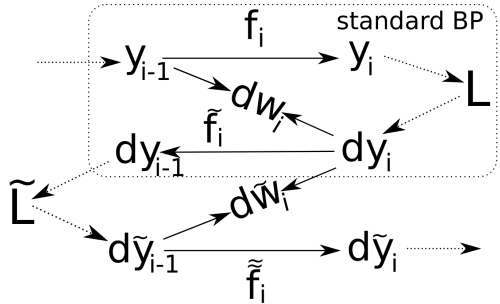}
\caption{The scheme represents three passes of Loss IBP algorithm. Two of them are the parts of standard backpropagation. It also shows which vectors are used for weight derivative computation.}
\label{fig:derivatives}
\end{figure}

\subsection{Loss IBP implementation}
\label{sec:lossimpl}

In this section we will show how to efficiently compute the weight gradients for the additional loss function (\ref{eq:ibp:fastloss}). To optimize $\tilde{L}(dy_0)$, we need to look at the backward pass from another point of view. We may consider that the derivatives $dy_K$ are the first layer of a \textit{reverse} neural network that has $dy_0$ as its output. Indeed, all transformation functions $f_i$ have reverse pairs $\tilde{f}_i$ that are used to propagate the derivatives \eqref{eq:jacobian}. If we consider these pairs as the original transformation functions, they have their own inverse pairs $\tilde{\tilde{f_i}}$. 

Therefore we consider the derivatives $dy_i$ as activations and the backward pass as a forward pass for the reverse network. As in standard backpropagation, after such a ``forward'' pass we compute the loss function $\tilde{L}(dy_0)$. The next step is quite natural: we need to initialize the input vector $y_0$ with the gradients $d\tilde{y}_0 = \nabla_{dy_0}\tilde{L}(dy_0)$ and perform another ``backward'' pass that has the same direction as the original forward pass. At the same time the derivatives with respect to the weights $d\tilde{w}_i = \nabla_{w_i} \tilde{L}(dy_0)$ must be computed. Fig.\ \ref{fig:derivatives} shows the general scheme of the derivative computation. The top part corresponds to the standard backpropagation procedure.

An important subset of transformation functions $f_i(y_{i-1}; w_i)$ is linear functions. It includes convolutional layers, fully connected layers, subsampling layers, and other types. In Section \ref{sec:theorems} we show that if a function $f_i$ is linear, i.e. $f_i(y_{i-1}; w_i) = y_{i-1} \cdot w_i$ then
\begin{enumerate}
\item $d\tilde{y}_i = d\tilde{y}_{i-1} \cdot w_i,$
\item $dw_i = y_{i-1}^T \cdot dy_i$, and $d\tilde{w}_i = d\tilde{y}_{i-1}^T \cdot dy_i$
\end{enumerate}

Therefore, in the case of a linear function $f_i$, we can propagate third pass activations the same way as we do on the first pass, i.e., multiplying them on the same matrix of weights $w_i$. This statement remains true for element-wise multiplication, as it can be considered as matrix multiplication as well. The weight derivatives $d\tilde{w}_i$ are also computed the same way as $dw_i$ in the standard BP algorithm. This fact allows us to easily implement Loss IBP using the same procedures as for standard BP.

Moreover, in Section \ref{sec:theorems} we also show that if the function $f_i(y_{i-1}; w_i)$ has a symmetric Jacobian $J_i(y_{i-1}; w_i)$, then $\tilde{\tilde{f}}_i(d\tilde{y}_{i-1}; w_i) = \tilde{f}_i(dy_i; w_i)$.
This property is useful for implementation of the non-linear functions. The summary of the Loss IBP algorithm is given in Algorithm \ref{alg:ibp}.

It is easy to compare the computation time for standard BP and Loss IBP. We know that convolution and matrix multiplication operations occupy almost all the processing time. As we see, IBP needs one more forward pass and one more calculation of weight gradients. If we assume that for each layer the forward pass, backward pass and calculation of derivatives all take approximately the same time, then IBP requires about $2/3 \approx 66\%$ more time to train the network. The experiments have shown that the additional time is about $50\%$. It is less than the approximated $66\%$, because both versions contain fixed time procedures such as batch composing, data augmentation, etc. At the same time Loss IBP is faster than Prediction IBP on approximately $20 \%$.

\section{Fast versions of Tangent BP and Adversarial Training}


\subsection{Fast Tangent BP}
\label{sec:fasttangent}

Let us change the additional loss function in Eq.\ \eqref{eq:tbpfun} such that we penalize the sensitivity of the main loss function $L(p(x))$ instead of predictions $p(x)$ themselves:
\begin{equation} \label{eq:fasttbploss}
R(x) = \tilde{L}(\nabla_\theta p(g(x,\theta))|_{\theta=0}) = \nabla_\theta p(g(x,\theta))|_{\theta=0} \cdot J_{K+1}^T = \nabla_\theta L(p(g(x,\theta)))|_{\theta=0}
\end{equation}
In this case the computations can be simplified. Notice, that
\[
\nabla_\theta L(p(g(x,\theta)))|_{\theta=0} = \prod_{i=K+1}^1 J_i(y_{i-1}) \cdot \nabla_\theta x = \nabla_x L(p(x)) \cdot \nabla_\theta x  = dy_0 \cdot \nabla_\theta x,
\]
Therefore $R(x)$ can be directly computed in the end of the backward pass by multiplying the gradient $dy_0$ on the tangent vector $\nabla_\theta x$. In Section \ref{sec:sup:tangent} we show that this modification of Tangent BP is equivalent to Loss IBP with the additional loss function $\tilde{L}(dy_0) = dy_0 \cdot \nabla_\theta x$ instead of $\tilde{L}(dy_0) = \frac{1}{r}||dy_0||_r^r$. Therefore, this version of Tangent BP can be implemented using $\approx 20\%$ less time than original Tangent BP. We refer to it as Fast TBP.

\subsection{Fast Adversarial Training}
\label{sec:fastat}

Using Taylor expansion for the loss of adversarial example $L(p(x^*(x;\epsilon)))$, we can get
\begin{equation} \label{eq:taylor}
L(p(x^*(x;\epsilon))) = L(p(x + \epsilon \, sign(\nabla_x L(p(x))))) = L(p(x)) + \epsilon ||\nabla_x L(p(x))||_1 + o(\epsilon)
\end{equation}
Combining (\ref{eq:atorigloss}) and (\ref{eq:taylor}), we can approximate $L_{min}(x)$ as
\begin{equation}
\begin{aligned}
L(p(x)) + \frac{\epsilon}{2} ||\nabla_x L(p(x))||_1 + o(\epsilon) \approx L(p(x+\frac{\epsilon}{2} \, sign(\nabla_x L(p(x))))) = L(p(x^*(x;\frac{\epsilon}{2})))
\end{aligned}
\end{equation}
It is easy to notice, that the usage of $L_{min}(x)$ instead of $L(p(x^*(x;\epsilon)))$ just scales the hyperparameter $\epsilon$, which needs to be tuned anyway. At the same time, the calculation of gradients $\nabla_{w_i} L(p(x))$ takes computation time. Therefore, the Adversarial Training algorithm can be sped up by avoiding the calculation of $\nabla_{w_i} L(p(x))$, and using only the gradients $\nabla_{w_i} L(p(x^*))$. Compared with the originally proposed loss $L_{min}(x)$, the optimal parameter $\epsilon$ must be $2$ times lower. Similar to Tangent BP, this trick also saves $\approx 20\%$.

Now we can see the difference between Loss IBP and Adversarial Training. While Loss IBP minimizes only the first derivative, and does not affect higher orders of the derivatives of the loss functions $L(p(x))$ such as curvature, Adversarial Training essentially minimizes all orders of the derivatives $\partial^n L(p(x)) / \partial^n x$ with the predefined weight coefficients between them. In the case of a highly nonlinear true data distribution $P(y|x)$ this might be a disadvantage. In Section \ref{sec:experiments} we show that none of these algorithms outperform another one in all the cases. 

\section{Experiments} \label{sec:experiments}

In the experimental part we compared all algorithms and their modifications in different aspects. We performed the experiments on two benchmark image classification datasets: MNIST (\cite{lecun1998gradient}) and CIFAR-10 (\cite{krizhevsky2009learning}) using the ConvNet toolbox for Matlab \footnote{https://github.com/sdemyanov/ConvNet}. In all experiments we used the following parameters: 1) the batch size $32$, 2) initial learning rate $\alpha = 0.1$, 3) momentum $m = 0.9$, 4) exponential decrease of the learning rate, i.e., $\alpha_t = \alpha_{t-1} \cdot \gamma$, 5) each convolutional layer was followed by a scaling layer with \textit{max} aggregation function among the region of size $3 \times 3$ and stride $2$, 6) \textit{relu} nonlinear functions on the internal layers, 7) final softmax layer combined with the negative log-likelihood loss function. We trained the classifiers for $80$ epochs with the coefficient $\gamma = 0.98$, so the final learning rate was $0.1 \cdot 0.98^{80} \approx 0.02$. For the experiments on MNIST we employed a network with two convolutional layers with $32$ filters of size $4 \times 4$ (padding $0$) and $64$ filters of size $5 \times 5$ (padding $2$) and one internal FC layer of length $256$. The experiments on CIFAR were performed on the network with $3$ convolutional layers with the filter size $5 \times 5$ (paddings $0$, $2$ and $2$), and one internal FC layer of length $256$.

In all our experiments we used $L_1$-norm additional loss function as we had found that it always works better that $L_2$-norm. For Tangent BP algorithm we used $5$ tangent vectors for each image in the training set, corresponding to $x$ and $y$ shifts, $x$ and $y$ scaling and rotation. The employed value of standard deviation for the Gaussian filter was $\sigma = 0.9$. For numerical stability reasons we omitted multiplication on softmax gradients on additional forward and backward passes in Prediction IBP and Original TBP algorithms.

\subsection{Classification accuracy}
\label{sec:accuracy}

First we compared the performance of all algorithms and their modifications. We trained the networks on $10$ different subsets of MNIST and CIFAR-10 datasets of size $10000$ with different initial weights and shuffling order. Each dataset was first normalized to have pixel values within $[0 \; 1]$ and then the mean pixel value was subtracted from all images. The results are presented in Table \ref{tbl:final1}.

\begin{table}[h]
\centering
\caption{Mean errors (\%), best parameters, and computation time of one epoch on MNIST and CIFAR-10 datasets for Standard backpropagation (BP), and two version of Invariant backpropagation (IBP), Adversarial Training (AT) and Tangent backpropagation (TBP) each.}
\label{tbl:final1}
\small
\begin{tabular}{l|ccc|ccc}
\hline
& \multicolumn{3}{c}{MNIST} & \multicolumn{3}{c}{CIFAR-10}\\
\hline
& Error, \% & Best $\beta$ or $\epsilon$ & Time, s & Error, \% & Best $\beta$ or $\epsilon$ & Time, s \\
\hline 
Standard BP & $1.21 \pm 0.08$ & N/A & $1.51$ & $34.7 \pm 0.6$ & N/A & $2.84$\\
Prediction IBP & $\textbf{0.90} \pm 0.10$ & $1.0$ & $2.64$ & $32.6 \pm 0.4$ & $0.1$ & $5.20$\\
Loss IBP & $1.09 \pm 0.11$ & $0.03$ & $2.25$ & $33.1 \pm 0.5$ & $0.003$ & $4.24$\\
Original AT & $\textbf{0.89} \pm 0.05$ & $0.05$ & $2.66$ & $34.7 \pm 0.3$ & $0.0003$ & $5.40$\\
Fast AT & $\textbf{0.89} \pm 0.06$ & $0.03$ & $2.28$ & $34.7 \pm 0.6$ & $0.0003$ & $4.78$\\
Original TBP & $1.07 \pm 0.12$ & $0.01$ & $7.47$ & $\textbf{27.2} \pm 0.7$ & $0.1$ & $15.55$\\
Fast TBP & $1.21 \pm 0.08$ & $0.0003$ & $5.38$ & $34.7 \pm 0.3$ & $0.0003$ & $10.30$\\
\end{tabular}
\end{table}

First, we can see that all algorithms except Fast TBP can decrease classification error compared with the standard BP. We suppose that the lack of improvement by Fast TBP can be explained by a weak connection between the behavior of the loss function $L(p(x))$ and predictions $p(x)$ themselves. While $L(p(x))$ is trained to be robust to predefined transformations, the predictions $p(x)$ might remain sensitive to them. Further we discuss only Original TBP.

Second, we can notice that Original and Fast AT demonstrate identical performance, thus confirming our suggestion about a possibility to speed up the algorithm. The achieved speed up is $17\%$ and $13\%$. We can also see that the best parameters of $\epsilon$ for MNIST datasets differ in $\approx$ 2 times, what was also predicted by our considerations. Further we do not differentiate between Original and Fast AT, and refer to them as AT.

Third, we can conclude that Prediction IBP shows better results than Loss IBP (improvement on $26\%$ vs $10\%$ on MNIST and $6.1\%$ vs $4.6\%$ on CIFAR), while being slightly slower (on $17\%$ and $23\%$ accordingly). Since Prediction IBP can be seen as a modification of Original TBP, while Loss IBP is equivalent to a modification of Fast TBP, the reason might also be a weak connection between $L(p(x))$ and $p(x)$.

Forth, we observe that the algorithms demonstrate different performance on MNIST and CIFAR-10 datasets. The best results on MNIST are achieved by Prediction IBP and AT, while the best result on CIFAR-10 is achieved by Tangent BP. Notice, that the improvement of Tangent BP on CIFAR-10 dataset ($22\%$) is much larger, than the next best result of Prediction IBP ($6.1\%$). At the same time, AT algorithm could not improve the accuracy at all, achieving the best accuracy using the lowest possible value of the parameter $\epsilon=0.0003$. However, the Tangent BP algorithm works much slower than the competitors. 

We suppose that such results can be explained by a high non-linearity of a decision function. As it was shown in Section \ref{sec:fastat}, AT minimizes not only the first order of the loss function derivatives, but also all other orders, thus preventing the classifier from learning such non-linearity. At the same time, Prediction IBP just makes the predictions less sensitive to variations in the input vector, specified by $\nabla_x L(p(x))$. In the case of highly non-linear decision function this might be not necessary. Unlike both AT and IBP, Tangent BP uses prior knowledge to train invariance in directions that the predictions must always be invariant to. This allows it to achieve the best performance on CIFAR-10.

\subsection{Robustness to Adversarial noise}
\label{sec:advnoise}

\begin{figure*}[h]
\begin{minipage}[b]{.45\linewidth}
  \centering
  \centerline{\includegraphics[width=8cm, height=4.5cm]{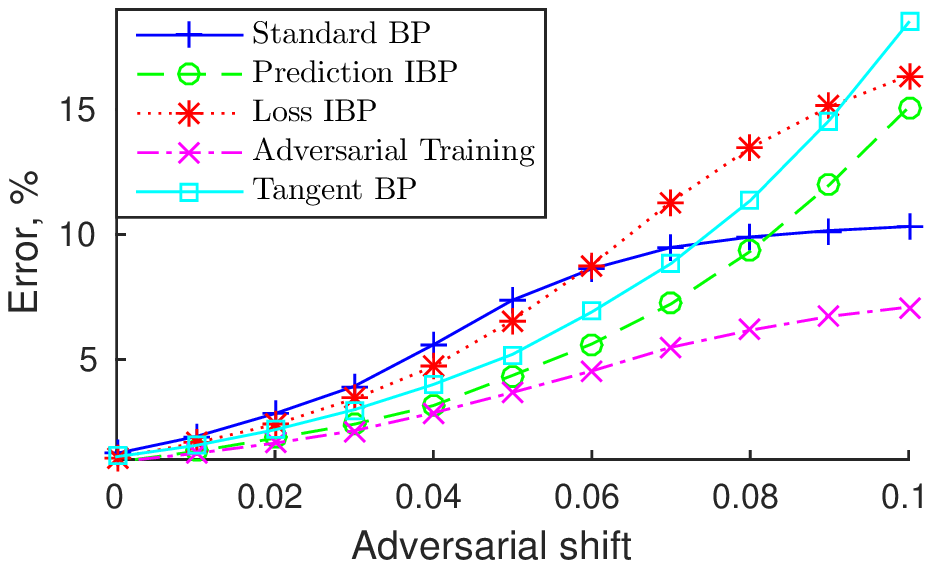}}
  \centerline{(a) MNIST dataset}\medskip
\end{minipage}
\hfill
\begin{minipage}[b]{0.45\linewidth}
  \centering
  \centerline{\includegraphics[width=8cm, height=4.5cm]{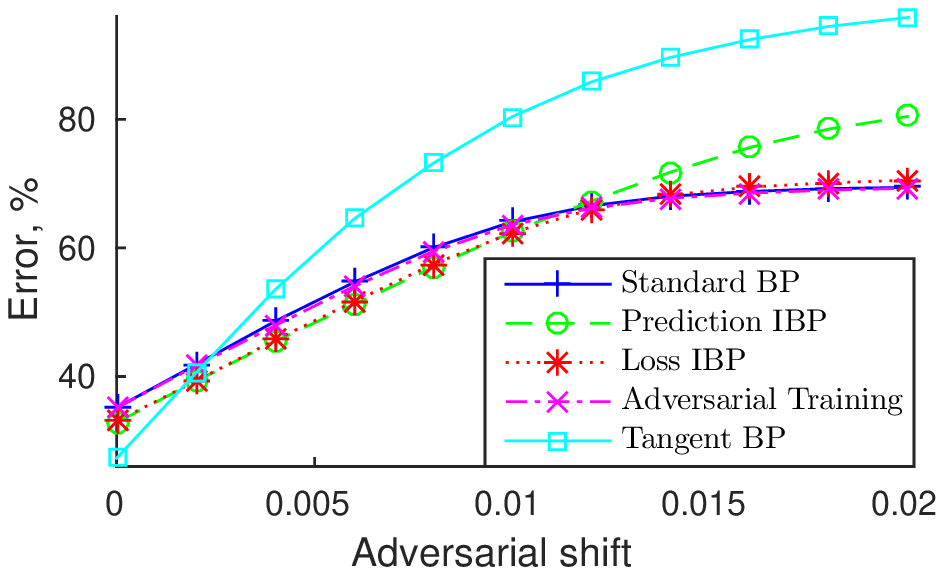}}
  \centerline{(b) CIFAR-10 dataset}\medskip  
\end{minipage}
\caption{Errors of competing algorithms on test sets, corrupted by different levels adversarial noise}
\label{fig:advnoise}
\end{figure*}

We next measured the sensitivity of all algorithms to adversarial noise. We employed the classifiers trained in Section \ref{sec:accuracy} with the parameters, which yield the best accuracy, and measured performance of the classifiers on the test sets, corrupted by adversarial noise. Adversarial examples were generated using Eq.\ \eqref{eq:advex}. The results are presented in Fig.\ \ref{fig:advnoise}, where we show the errors at the variation of $\epsilon$. It is important to keep in mind that performance of the classifiers significantly depends on the value of a regularization parameter.

Firstly notice that CIFAR-10 classifiers are much more sensitive to adversarial noise than those trained on MNIST dataset. As expected, the most robust classifier was trained by Adversarial Training algorithm. It is the only one which constantly remains better than standard BP classifier. Other classifiers show better results until a certain point, when the level of noise becomes too high. Interestingly, while Tangent BP demonstrates the best results on CIFAR-10 dataset, its performance degrades much faster than the performance of other classifiers on both MNIST and CIFAR-10. Note, that despite the ratio of best $\beta$ values for Prediction IBP and Loss IBP is the same in both cases, they demonstrate different behavior.

\subsection{Robustness to Gaussian noise}
\label{sec:gausnoise}

\begin{figure*}[h]
\begin{minipage}[b]{.45\linewidth}
  \centering
  \centerline{\includegraphics[width=8cm, height=4.5cm]{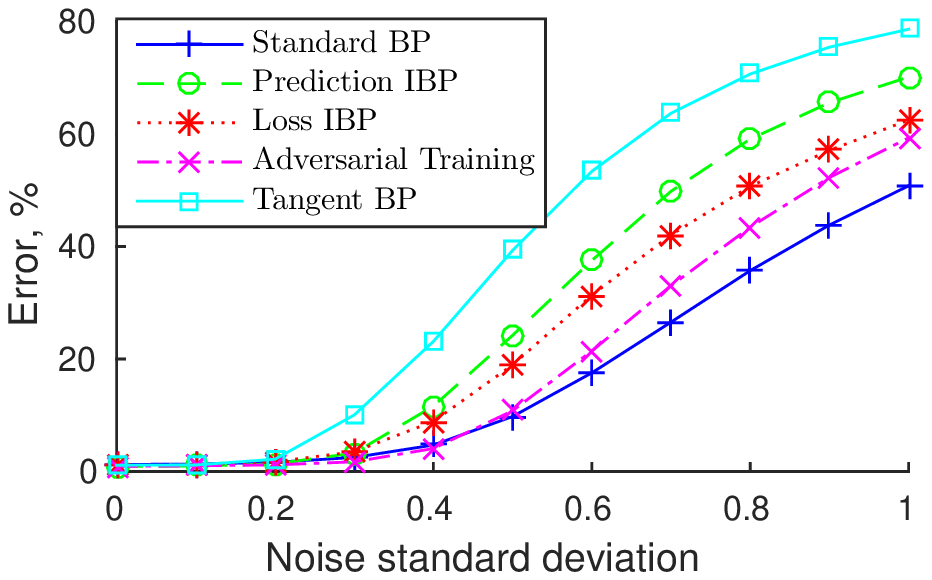}}
  \centerline{(a) MNIST dataset}\medskip
\end{minipage}
\hfill
\begin{minipage}[b]{0.45\linewidth}
  \centering
  \centerline{\includegraphics[width=8cm, height=4.5cm]{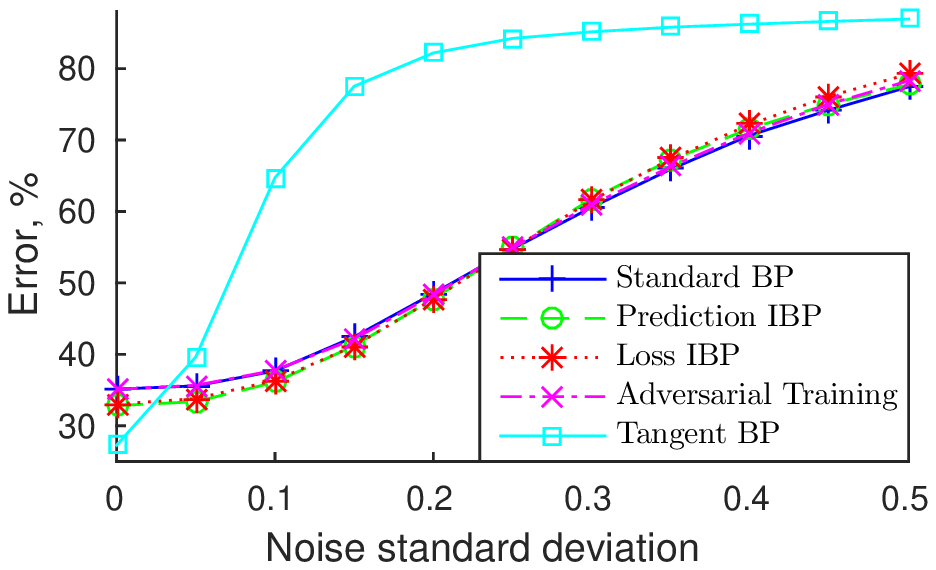}}
  \centerline{(b) CIFAR-10 dataset}\medskip  
\end{minipage}
\caption{Errors of competing algorithms on test sets, corrupted by different levels Gaussian noise}
\label{fig:gaussnoise}
\end{figure*}

After that we measured the sensitivity of the same classifiers to Gaussian noise. The results are presented in Fig.\ \ref{fig:gaussnoise}. 
Surprisingly, the most robust classifier on MNIST dataset was trained by standard BP. We thus see that robustness to adversarial noise and other predefined transformations makes a classifier more sensitive to Gaussian noise. At the same time, Tangent BP classifier remains the most sensitive to Gaussian noise as well. On CIFAR-10 dataset it is the only classifier which degrades significantly faster than others.

\subsection{Dataset size and Data augmentation}

\begin{figure*}[h]
\begin{minipage}[b]{.45\linewidth}
  \centering
  \centerline{\includegraphics[width=8cm, height=4cm]{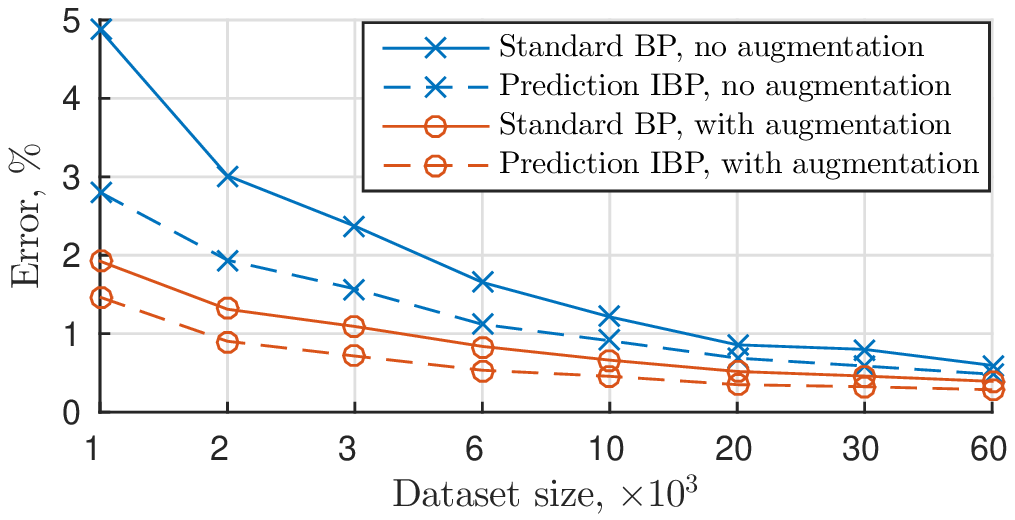}}
  \centerline{(a) Classification errors}\medskip
\end{minipage}
\hfill
\begin{minipage}[b]{0.45\linewidth}
  \centering
  \centerline{\includegraphics[width=8cm, height=4cm]{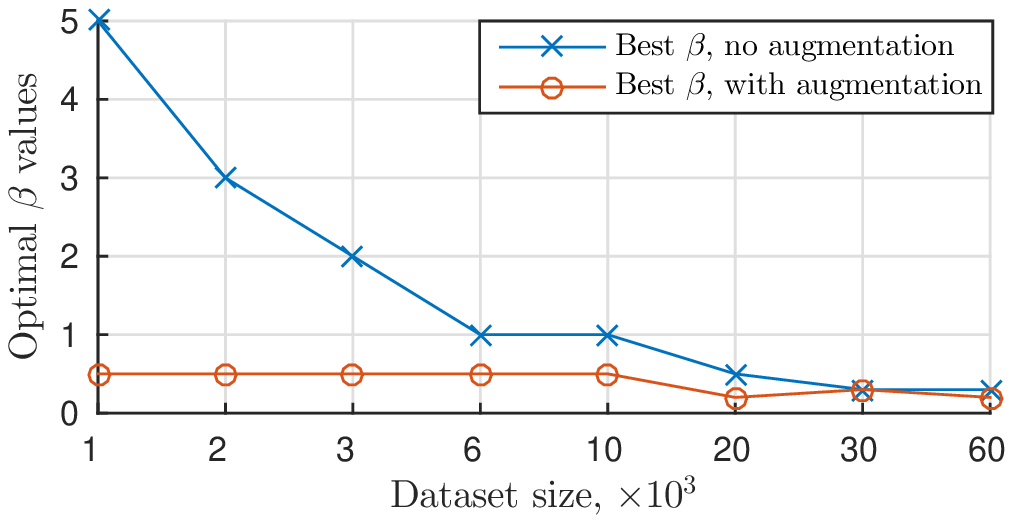}}
  \centerline{(b) Optimal $\beta$ values}\medskip  
\end{minipage}
\caption{Performance of standard BP and Prediction IBP on different size subsets of MNIST dataset with and without data augmentation. The optimal $\beta$ values are provided on the right plot.}
\label{fig:dataaug}
\end{figure*}

We have also established how the dataset size and data augmentation affects the Prediction IBP improvement. We performed these experiments on subsets of the MNIST dataset using the same parameters as in Section \ref{sec:accuracy}. In data augmentation regime we randomly modified each training object every time it was accessed according to the following parameters: 1) range of shift from the central position in each dimension - $[-2,\; 2]$ pixels, 3) range of scaling in each dimension - $[0.7,\; 1.4]$, 3) range of rotation angle - $[-18,\; 18]$ degrees, 4) pixel value if the pixel is out of the original image - $0$. In order to decrease the variance we trained the networks for $100$ epochs without data augmentation and for $150$ epochs with it. 

The results are summarized in Fig.\ \ref{fig:dataaug}. We see that without data augmentation smaller datasets require more regularization, i.e., larger $\beta$. The relative improvement is also higher: it is $43\%$ for $1k$ samples and $18\%$ for $60k$. We thus see that the larger the dataset is, the less the network overfits, and the less improvement we can obtain from regularization. With data augmentation the improvement of IBP is less, but does not converge to $0$ even when the full training set is used. Interestingly, the optimal value of $\beta$ remains approximately on the same level for all dataset sizes. Therefore we can conclude that data augmentation cannot completely substitute IBP regularization as the last one enforces robustness to variations, which are not represented by additionally generated objects.

\section{Conclusion}
We proposed two versions of the Invariant Backpropagation algorithm, which extends the standard Backpropagation in order to enforce robustness of a classifier to variations in the input vector. While Loss IBP trains the main loss function to be insensitive to any variations, Prediction IBP trains the predictions to be insensitive to variations in the direction of the gradient $\nabla_x L(p(x))$. We have demonstrated that the weight gradients for Loss IBP can be efficiently computed using only one additional forward pass, which is identical to the original forward pass for the majority of layer types. We experimentally established that Prediction IBP achieves higher classification accuracy on both MNIST and CIFAR-10 datasets, but requires $\approx 20\%$ more time than Loss IBP. Additionally we proposed fast versions for both Tangent BP and Adversarial Training algorithms. While the fast version of Tangent BP does not improve classification accuracy, the modification of Adversarial Training algorithm demonstrates the same performance as the originally proposed algorithm, being $\approx 15\%$ faster.

In the experimental part we performed comparison of all algorithms and their modifications in terms of classification accuracy and robustness to noise. We have found that none of the algorithms outperforms others in all cases. While the best results on MNIST are achieved by Prediction IBP and Adversarial Training, Tangent BP significantly outperformed others on CIFAR-10. At the same time Tangent BP classifier is the most sensitive to Gaussian and Adversarial noise on both datasets. Additionally we demonstrated that the regularization effect of Prediction IBP remains visible even on the full size MNIST dataset with data augmentation, so the methods can be applied together. The choice of a particular regularizer depends on the properties of a dataset.

\bibliography{references}
\bibliographystyle{iclr2016_conference}

\section{Supplementary material}

\subsection{Reverse function theorems}
\label{sec:theorems}

First, let us notice that the forward and backward passes of Loss IBP are performed in the same way as in the standard backpropagation algorithm. Then the additional loss function \eqref{eq:ibp:fastloss} is computed, and its derivatives are used as input for the propagation on the third pass. As it follows from \eqref{eq:ibp:fastloss}, for $r=2$ the gradients are 
\[
d\tilde{y}_0 = \frac{1}{2} \frac{\partial ||dy_0||_2^2}{\partial (dy_0)} = dy_0,
\]
i.e., coincide with the gradients $dy_0 = \nabla_x L(p(x))$. For $p=1$, they are the \textit{signs} of $dy_0$:
\[
d\tilde{y}_0 = \frac{\partial ||dy_0||_1}{\partial (dy_0)} = sign(dy_0)
\]
In Section \ref{sec:lossimpl} we described double reverse functions $\tilde{\tilde{f}}(d\tilde{y}_{i-1}; w_i)$. Let us additionally introduce functions $g_i$ and their reverse pairs $\tilde{g}_i$ as
\[
dw_i = g_i(y_{i-1}, dy_i), \text{ and }
d\tilde{w}_i = \tilde{g}_i(d\tilde{y}_{i-1}, dy_i)
\]
Now we can prove the following theorems.
\begin{theorem} \label{th:linear}
Let us assume that $f_i$ is linear, i.e., $f_i(y_{i-1}; w_i) = y_{i-1} \cdot w_i$, where matrix multiplication is used. Then 
\begin{enumerate}
\item $\tilde{\tilde{f_i}} = f_i$, i.e., $d\tilde{y}_i = \tilde{\tilde{f_i}}(d\tilde{y}_{i-1}; w) = d\tilde{y}_{i-1} \cdot w_i,$
\item $\tilde{g_i} = g_i$, i.e., $dw_i = y_{i-1}^T \cdot dy_i$, and $d\tilde{w}_i = d\tilde{y}_{i-1}^T \cdot dy_i$
\end{enumerate}
\end{theorem} \vspace{-1em}

\begin{proof}
First, notice that the reverse of any function is always linear:
\begin{equation} \label{eq:ibp:any_inverse}
dy_{i-1} = \tilde{f_i}(dy_i; w_i) = dy_i \cdot J_i(y_{i-1}; w_i)
\end{equation}
In the case of a linear function $f_i$ the reverse function $\tilde{f_i}$ is known: 
\begin{equation} \label{eq:ibp:inverse}
dy_{i-1} = dy_i \cdot J_i(y_{i-1}; w_i) = dy_i \cdot w_i^T
\end{equation}
Now let us consider the double reverse functions $\tilde{\tilde{f}}(d\tilde{y}; w)$, such that $d\tilde{y}_i = \tilde{\tilde{f_i}}(d\tilde{y}_{i-1}; w_i)$. Compared with linear $f$, its reverse function $\tilde{f}$ multiplies its first argument on the transposed parameter. The same is true for the double reverse function $\tilde{\tilde{f}}$ compared with $\tilde{f}$, i.e.: 
\begin{align*}
d\tilde{y}_i = \tilde{\tilde{f_i}}(d\tilde{y}_{i-1}; w_i) = d\tilde{y}_{i-1} \cdot (w_i^T)^T = d\tilde{y}_{i-1} \cdot w_i
\end{align*}
This proves the first statement.

Next, in the case of linear function $f_i$ we also know the function $g_i(y_{i-1}, dy_i)$ which computes the weight derivatives $dw_i$ \eqref{eq:weights_der}: 
\begin{equation} \label{eq:ibp:invder}
dw_i = g_i(y_{i-1}, dy_i) = y_{i-1}^T \cdot dy_i.
\end{equation}
Let us again consider the backward pass $\tilde{f_i}$ as the forward pass for the reverse net. Since the function $\tilde{f_i}$ is linear, the formula for derivative calculation of reverse net is also \eqref{eq:ibp:invder}. However, as it follows from \eqref{eq:ibp:inverse} the reverse net uses the \textit{transposed} matrix of weights for forward propagation, so the result of the derivative calculation is also transposed with respect to the matrix $w_i$. Also note that since $dy_i$ acts as activations in the reverse net, we pass it as the first argument, and $d\tilde{y}_{i-1}$ as the second. Therefore,
\begin{equation} \label{eq:ibp:add_weights_der}
d\tilde{w}_i = g_i(dy_i, d\tilde{y}_{i-1})^T = (dy_i^T \cdot d\tilde{y}_{i-1})^T = d\tilde{y}_{i-1}^T \cdot dy_i,
\end{equation}
and this proves the part 2.
\end{proof}

\begin{theorem} \label{th:linear2}
If the function $f_i(y_{i-1}; w_i)$ has a symmetric Jacobian $J_i(y_{i-1}; w_i)$, then $\tilde{\tilde{f}}_i(d\tilde{y}_{i-1}; w_i) = \tilde{f}_i(dy_i; w_i)$.
\end{theorem} \vspace{-1em}
\begin{proof}

Indeed, from \eqref{eq:ibp:any_inverse} we know that any reverse function is linear, and its argument is multiplied on the Jacobian  $J_i(y_{i-1}; w_i)$. From Theorem \ref{th:linear} we also know that the reverse for a linear function multiplies its argument of the \textit{transposed} set of weights, i.e. $d\tilde{y}_i = d\tilde{y}_{i-1} \cdot J_i^T(y_{i-1}; w_i)$. Therefore, if the Jacobian is symmetric, then $\tilde{\tilde{f}}_i(d\tilde{y}_{i-1}; w_i) = \tilde{f}_i(dy_i; w_i)$.
\end{proof}

\subsection{Implementation of particular layer types}

\paragraph{A fully connected layer} is a standard linear layer, which transforms its input by multiplication on the matrix of weights: $y_{i} = y_{i-1} \cdot w_i + b_i$, where $b_i$ is the vector of biases. Notice that on the backward pass we do not add any bias to propagate the derivatives, so we do not add it on the third pass as well and do not compute additional bias derivatives. This is the difference between the first and the third passes. If \textbf{dropout} is used, the third pass should use the same dropout matrix as used on the first pass.

\paragraph{Non-linear activation functions} can be considered as a separate layer, even if they are usually implemented as a part of each layer of the other type. They do not contain weights, so we write just $f(z)$. The most common functions are: (i) sigmoid, $f(z) = 1 / (1 + e^{-z})$, (ii) rectified linear unit (\textit{relu}), $f(z) = \max(z, 0)$, and (iii) softmax, $f(z_i) = e^{z_i} / \sum_j e^{z_j}$. All of them are differentiable (except \textit{relu} in $0$, but it does not cause uncertainty) and have a symmetric Jacobian matrix, so according to Theorem \ref{th:linear2} the third pass is the same the backward pass. For example, in the case of the \textit{relu} function this means that $d\tilde{y}_i = d\tilde{y}_{i-1} * I(y_{i-1} > 0)$, where element-wise multiplication is used.

\paragraph{Convolution layers} perform 2D filtering of the activation maps with the matrices of weights. Since each element of $y_i$ is a linear combination of elements of $y_{i-1}$, convolution is also a linear transformation. Linearity immediately gives that $\tilde{\tilde{f_i}}(d\tilde{y}_{i-1}; w_i) = f_i(y_{i-1}; w_i)$ and $d\tilde{w}_i = d\tilde{y}_{i-1}^T \cdot dy_i$. Therefore the third pass of convolutional layer repeats its first pass, i.e., it is performed by convolving $d\tilde{y}_{i-1}$ with the same filters using the same stride and padding. As with the fully connected layers, we do not add biases to the resulting maps and do not compute their derivatives.

\paragraph{The scaling layer} aggregates the values over a region to a single value. Typical aggregation functions $f_i(y_{i-1})$ are \textit{mean} and \textit{max}. As it follows from their definition, both of them also perform linear transformations, so $d\tilde{y}_i = f_i(d\tilde{y}_{i-1})$. Notice that in the case of the \textit{max} function it means that on the third pass the same elements of $d\tilde{y}_{i-1}$ should be chosen for propagation to $d\tilde{y}_i$ as on the first pass regardless of what value they have.

\begin{algorithm}[h]
\caption{Invariant backpropagation: a single batch processing description} \label{alg:ibp}
\begin{enumerate}
\item Perform standard forward and backward passes, and compute the derivatives $dw$ for the main loss function.
\item Perform additional forward pass using the derivatives $dy_0$ or signs $sign(dy_0)$ as activations. On this pass:
\begin{itemize}
\item do not add biases to activations
\item use backward versions of non-linear functions
\item on max-pooling layers propagate the same positions as on the first pass
\end{itemize}
\item Compute the derivatives $d\tilde{w}$ for the additional loss function $\tilde{L}$ the same way as $dw$. Initialize the bias derivatives $d\tilde{w}$ to $0$.
\item Update the weights according to Eq.\ \ref{eq:ibp:invupdaterule}.
\end{enumerate}
\end{algorithm}

\subsection{Regularization properties of Loss IBP}

In the case of $\tilde{L}_2$ loss function \eqref{eq:ibp:fastloss}, we can derive some interesting theoretical properties. Using the Cauchy-Schwarz inequality, we can obtain, that
\[
||\nabla_x L||_2^2 \leq ||\nabla_x y_K||_2^2 \cdot ||\nabla_{y_K} L||_2^2 \leq ||\nabla_x y_{K-1}||_2^2 \cdot ||\nabla_{y_{K-1}} y_K||_2^2 \cdot ||\nabla_{y_K} L||_2^2
\]
The most common loss functions for the predictions $p(x)=y_K$ and true labels $l(x)$ are the squared loss $L(p(x)) = \frac{1}{2} \sum_{i=1}^M (p_i(x)-l_i(x))^2$ and the cross-entropy loss $-\sum_{i=1}^M l_i(x) \log p_i(x)$, applied to the softmax output layer $p_i(x) = \phi(z_i) = e^{z_i} / \sum_{j=1}^M e^{z_j}$. Here $M$ is the number of neurons in the output layer (number of classes), and $z = y_{K-1}$. In the first case we have $\nabla_{y_K} L = p(x) - l(x)$, in the second case we can show that $\nabla_{y_{K-1}} L = p(x) - l(x)$. Therefore, the strength of $\tilde{L}_2$-function Loss IBP regularization decreases when the predictions $p(x)$ approach the true labels $l(x)$. This property prevents overregularization when the classifier achieves high accuracy. Notice, that if a network has no hidden layers, then $\nabla_x y_{K-2} = w$, i.e., in this case $||\nabla_x L||_2^2$ penalty term can be considered as a weight decay regularizer, multiplied on $p(x) - l(x)$.

For the model of a single neuron we can derive another interesting property. In (\cite{bishop1995training}) it was demonstrated that for a single neuron with the $L_2$-norm loss function noise injection is equivalent to the weight decay $||w||_2^2$ regularization. In Section \ref{sec:noiseinj} we show, that \textbf{if the negative log-loss function is used, noise injection becomes equivalent to the Loss IBP regularizer}.

\subsection{Noise injection} \label{sec:noiseinj}

Assuming Gaussian noise $\mu \sim N(0,\sigma^2 I)$, such that $E[\mu]=0$ and $E[\mu^T \mu] = \sigma^2 I$, we can get approximate an arbitrary loss function $L(p(x))$ as
\[
E[L(p(x+\mu))] \approx E \left[L(p(x)) + \nabla_x L(p(x)) \mu^T + \frac{1}{2}\mu H(x) \mu^T \right] = L(p(x)) + \frac{\sigma^2}{2} Tr(H(x)),
\]
where $Tr(H(x))$ is the trace of the Hessian matrix $H$, consisting of the second derivatives of $L(p(x))$ with respect to the elements of $x$. Solving the differential equation 
\[
Tr(H(x)) = \sum_{i=1}^N \frac{\partial^2 L}{\partial x_i^2} = \sum_{i=1}^N \left(\frac{\partial L}{\partial x_i}\right)^2 = ||\nabla_x L||_2^2,
\]
for each $x_i$ independently, we can find the following solution:
\[
L = - \left[ l(x) \ln |\sum_{i=1}^N x_i w_i + b| + (1-l(x)) \ln |1-\sum_{i=1}^N x_i w_i - b | \right],
\]
where $l = l(x) \in \{0,1\}$ is the class label for the object $x$. Indeed, assuming $p = \sum_{i=1}^N x_i w_i + b$, we obtain the first derivatives:
\begin{equation} \label{eq:firstder}
\left(\frac{\partial L}{\partial x_i}\right)^2 = \left[ l \frac{\pm w_i}{\pm p} + (1-l) \frac{\mp w_i}{\pm (1-p)}\right]^2 = w_i^2 \left( \frac{p - l}{p(1-p)}\right)^2 = w_i^2 \frac{p^2 - 2pl + l^2}{p^2(1-p)^2}
\end{equation}
Now we can compute the second derivatives:
\begin{equation} \label{eq:secondder}
\frac{\partial^2 L}{\partial x_i^2} = \frac{\partial}{\partial x_i} \left[ w_i \frac{p - l}{p(1-p)} \right] = w_i^2 \frac{p^2 - 2pl + l}{p^2 (1-p)^2}
\end{equation}
Notice, that the last expression uses $l$ instead of $l^2$. However if $l \in \{0, 1 \}$, then $l = l^2$, so the expressions \eqref{eq:firstder} and \eqref{eq:secondder} are equal. Therefore, when the negative log-likelihood function $L$ is applied to a single neuron without a non-linear transfer function, the Gaussian noise, added to the input vector $x$, is equivalent to the IBP regularization term $||\nabla_x L||_2^2$. This result is supported by the discussion in \cite{fawzi2015fundamental}, where the authors show that for the linear classifier the robustness to adversarial examples is bounded from below by the robustness to random noise. However, since $Tr(H(x))$ is only the expected value, the quality of approximation also depends on the number of iterations.

\subsection{Equivalence of Loss IBP and Fast TBP} \label{sec:sup:tangent}
In Section \ref{sec:fasttangent} we showed that the gradient $\nabla_\theta L(p(g(x,\theta)))|_{\theta=0}$  can be efficiently computed by multiplying the gradient $dy_0=\nabla_x L(p(x))$, obtained at the end of the backward pass, on the tangent vector $\nabla_\theta x$. We can demonstrate that Loss IBP with the additional loss function $\tilde{L}(dy_0) = dy_0 \cdot \nabla_\theta x$ is equivalent to Fast Tangent BP with the additional loss function \eqref{eq:fasttbploss}.

In Fast Tangent BP we perform an additional iteration of backpropagation through the linearized network, applied to a tangent vector $\nabla_\theta x$. The additional forward pass computes the following values:
\[
\tilde{y}_i = \nabla_\theta x^T \cdot \prod_{j=1}^i J_j^T, \,\,\, R(x) = \tilde{y}_{K+1}
\]
On the additional backward pass the computed gradients $d\tilde{y}_i = \partial R(x) / \partial \tilde{y}_i$ are therefore
\[
d\tilde{y}_i = \frac{\partial R(x)}{\partial \tilde{y}_{K}} \cdot \prod_{j=K}^{i+1} J_j =\prod_{j=K+1}^{i+1} J_j
\]
According to \eqref{eq:weights_der}, the weight gradients are then
\[
d\tilde{w}_i = \tilde{y}_{i-1}^T \cdot d\tilde{y}_i = \left(\nabla_\theta x^T \cdot \prod_{j=1}^{i-1} J_j^T \right)^T \cdot \prod_{j=K+1}^{i+1} J_j = \prod_{j=i-1}^1 J_j \cdot \nabla_\theta x \cdot \prod_{j=K+1}^{i+1} J_j
\]
We thus see that in order to compute additional weight derivatives $d\tilde{w}_i$, we need to compute the cumulative Jacobian products from both sides of the network. 

Let us now compute the same gradients $d\tilde{w}_i$ for Loss IBP with $\tilde{L}(dy_0) = dy_0 \cdot \nabla_\theta x$. In this case we initialize the third pass by the tangent vector $\nabla_\theta x^T = \partial \tilde{L}(dy_0) / \partial (dy_0)$. Thus the third pass values are
\[
d\tilde{y}_i = \nabla_\theta x^T \cdot \prod_{j=1}^{i} J_j^T
\]
According to \eqref{eq:ibp:add_weights_der}, the gradients are
\[
d\tilde{w}_i = d\tilde{y}_{i-1}^T \cdot dy_i = \left( \nabla_\theta x^T \cdot \prod_{j=1}^{i-1} J_j^T \right)^T \cdot \prod_{j=K+1}^{i+1} J_j = \prod_{j=i-1}^1 J_j \cdot \nabla_\theta x \cdot \prod_{j=K+1}^{i+1} J_j
\]
Therefore, the weight gradients of both algorithms are the same, so the algorithms are equivalent. 

\end{document}